\def\year{2017}\relax
\documentclass[letterpaper]{article}
\usepackage{aaai17}
\usepackage{times}
\usepackage{helvet}
\usepackage{courier}

\usepackage[utf8]{inputenc} 
\usepackage[T1]{fontenc}    
\usepackage{hyperref,color}       
\usepackage{url}            
\usepackage{booktabs}       
\usepackage{amsfonts}       
\usepackage{nicefrac}       
\usepackage{microtype}      

\usepackage{array}
\usepackage{amssymb,amsmath}
\usepackage{graphicx} 
\usepackage{subfigure} 
\usepackage{multirow}

\begin{document}
	
	\title{Unsupervised Learning with Truncated Gaussian Graphical Models}
	\author{Qinliang Su, Xuejun Liao,  Chunyuan Li, Zhe Gan and Lawrence Carin\\
		Department of Electrical \& Computer Engineering \\
		Duke University\\
		Durham, NC 27708-0291\\
	}
	
	\maketitle
	
	\begin{abstract}
		Gaussian graphical models (GGMs) are widely used for statistical modeling, because of ease of inference and the ubiquitous use of the normal distribution in practical approximations. However, they are also known for their limited modeling abilities, due to the Gaussian assumption. In this paper, we introduce a novel variant of GGMs, which relaxes the Gaussian restriction and yet admits efficient inference. Specifically, we impose a bipartite structure on the GGM and govern the hidden variables by {\em truncated} normal distributions. The nonlinearity of the model is revealed by its connection to rectified linear unit (ReLU) neural networks. Meanwhile, thanks to the bipartite structure and appealing properties of truncated normals, we are able to train the models efficiently using contrastive divergence. We consider three output constructs, accounting for real-valued, binary and count data. We further extend the model to deep constructions and show that deep models can be used for unsupervised pre-training of rectifier neural networks. Extensive experimental results are provided to validate the proposed models and demonstrate their superiority over competing models.
		
	\end{abstract}
	
	\section{Introduction}
	
	Gaussian graphical models (GGMs) have been widely used in practical applications \cite{honorio2009sparse,liu2013learning,meng2014learning,oh2014inference,su2015distributed,su2015convergence} to discover statistical relations of random variables from empirical data. The popularity of GGMs is largely attributed to the ubiquitous use of normal-distribution approximations in practice, as well as the ease of inference due to the appealing properties of multivariate normal distributions. On the downside, however, the Gaussian assumption prevents GGMs from being applied to more complex tasks, for which the underlying statistical relations are inherently non-Gaussian and nonlinear. It is true for many models that, by adding hidden variables and integrating them out, a more expressive distribution can be obtained about the visible variables; such models include Boltzmann machines (BMs) \cite{ackley1985learning}, restricted BMs (RBMs) \cite{hinton2002training,hinton2006fast,salakhutdinov2009deep}, and sigmoid belief networks (SBNs) \cite{neal1992connectionist}. Unfortunately, this approach does not work for GGMs since the marginal distribution of visible variables always remains Gaussian no matter how many hidden variables are added.
	
	Many efforts have been devoted to enhancing the representational versatility of GGMs. In \cite{frey1997continuous,frey1999variational}, nonlinear Gaussian belief networks were proposed, with explicit nonlinear transformations applied on random variables to obtain nonlinearity. More recently, \cite{su2016learning} proposed to employ truncated Gaussian hidden variables to implicitly introduce nonlinearity. An important advantage of truncation over transformation is that many nice properties of GGMs are preserved, which can be exploited to facilitate inference of the model. However, the models all have a directed graphical structure, for which it is difficult to estimate the posteriors of hidden variables due to the ``explaining away'' effect inherent in directed graphical models. As a result, mean-field variational Bayesian (VB) analysis was used. It is well known that, apart from the scalability issue, the independence assumption in mean-field VB is often too restrictive to capture the actual statistical relations. Moreover, \cite{su2016learning} is primarily  targeted at supervised learning. We notice that there are also other means of introducing nonlinearities into GGM \cite{radosavljevic2014neural,elidan2010copula}, but they are out of the scope of this paper.
	
	We consider an {\em{undirected}} GGM with truncated hidden variables. This serves as a counterpart of the directed model in \cite{su2016learning}, and it is particularly useful for unsupervised learning. Conditional dependencies are encoded in the graph structure of undirected graphical models. We impose a bipartite structure on the graph, such that it contains two layers (one hidden and one visible) and only has inter-layer connections, leading to a model termed a restricted truncated GGM (RTGGM). In RTGGM, visible variables are conditionally independent given the hidden variables, and vice versa. By exploiting the conditional independencies as well as the appealing properties of truncated normals, we show that the model can be trained efficiently using contrastive divergence (CD) \cite{hinton2002training}. This makes a striking contrast to the directed model in \cite{su2016learning}, where the conditionally-independent properties do not exist and inference is done based on mean-field VB approximation. 
	
	Although the variables in an RTGGM are conditionally independent, their marginal distributions are flexible enough to model many interesting data. Truncated real observations (e.g., nonnegative) are naturally handled by the RTGGM. We also develop three variants of the basic RTGGM, appropriate for modeling real, binary or count data. It is shown that all variants can also be trained efficiently by the CD algorithm. Furthermore, we extend two-layer RTGGMs to deep models, by stacking multiple RTGGMs together, and show that the deep models can be trained in a layer-wise manner. To evaluate the performance of the proposed models, we have also developed methods to estimate their partition functions, based on annealed importance sampling (AIS) \cite{salakhutdinov2008quantitative,neal2001annealed}. Extensive experimental results are provided to validate the advantages of the RTGGM models.
	
	\section{Related Work}
	The proposed RTGGM is a new member of the GGM family, and it is also closely related to the RBM \cite{hinton2002training}. One of the main differences between the two models is their inherent nonlinearities. In an RTGGM, the visible and hidden variables are related through smoothed ReLU functions, while they are related by sigmoid functions in an RBM. The ReLU is used extensively in neural networks and has achieved tremendous success due to its training properties \cite{jarrett2009best}. In light of this, there have been many efforts devoted to bringing the ReLU into the RBM formalism. For example, \cite{nair2010rectified} proposed to replace binary hidden units with a rectified Gaussian approximation. Although an ReLU-like nonlinearity is induced, the proposed model is only specified by two conditional distributions, while lacking an appropriately defined joint distribution. On the other hand, \cite{ravanbakhsh2016stochastic} proposed to use Exponential Family Harmoniums (EFH) \cite{welling2004exponential} and Bregman divergence to incorporate different monotonic nonlinearities into the RBM. The model preserves a joint-distribution description, but their conditional distributions are complicated and do not admit exact and efficient sampling. To overcome this, they need to approximate the conditional distributions as Gaussian and then sample from the approximate distributions. In contrast to the above models, the proposed RTGGM not only maintains an explicit joint distribution, but also preserves simple conditional distributions (truncated normals), allowing exact and efficient sampling. Moreover, because of the explicit joint distribution and the easily-sampled conditional distributions, we are able to estimate the partition function of the RTGGM, for performance evaluation. However, it is not clear how to estimate the partition function for models in \cite{nair2010rectified,ravanbakhsh2016stochastic}. Interestingly, we also note that the smoothed ReLU associated with the proposed RTGGM share some similarities with the leaky ReLU \cite{he2015delving}, as both have small nonzero slopes for negative inputs. 
	
	\section{Formulation of Restricted-Truncated Gaussian Graphical Models}
	\label{gen_inst}
	
	\subsection{Basic Model}
	Let ${\mathbf{x}}\in {\mathbb{R}}^n$ and ${\mathbf{h}}\in {\mathbb{R}}^m$ denote the visible and hidden variables, respectively. The joint probability distribution of an RTGGM is defined as
	\begin{equation} \label{RTGGM_joint_pdf_org}
		p({\mathbf{x}}, {\mathbf{h}}; {\boldsymbol{\Theta}}) = \frac{1}{Z} e^{- E({\mathbf{x}}, {\mathbf{h}})}{\mathbb{I}}({\mathbf{x}} \ge 0) {\mathbb{I}}({\mathbf{h}} \ge 0),
	\end{equation}
	where ${\mathbb{I}}(\cdot)$ is the indicator function, $E({\mathbf{x}}, {\mathbf{h}})$ is an energy function defined as
	\begin{align} \label{energy_fun}
		E({\mathbf{x}}, {\mathbf{h}}) &\triangleq \frac{1}{2}\left({\mathbf{x}}^T\text{diag}({\mathbf{a}})\,{\mathbf{x}} + {\mathbf{h}}^T\text{diag}({\mathbf{d}})\,{\mathbf{h}}  \right. \nonumber \\
		& \quad \left. - 2{\mathbf{x}}^T{\mathbf{W}} {\mathbf{h}} - 2{\mathbf{b}}^T{\mathbf{x}} - 2{\mathbf{c}}^T{\mathbf{h}} \right),
	\end{align}
	$Z$ is the partition function, the superscripted $T$ denotes matrix transpose, and ${\boldsymbol{\Theta}} \triangleq \{{\mathbf{W}}, {\mathbf{a}}, {\mathbf{d}}, {\mathbf{b}}, {\mathbf{c}}\}$ collects all model parameters. The joint distribution in \eqref{RTGGM_joint_pdf_org} can be equivalently written as
	\begin{equation} \label{RTGGM_joint_pdf}
		p({\mathbf{x}}, {\mathbf{h}}; {\boldsymbol{\Theta}}) = {\mathcal{N}}_T\left([{\mathbf{x}}^T, {\mathbf{h}}^T]^T \left| {\boldsymbol{\mu}},  {\mathbf{P}}^{-1}\right.\right),
	\end{equation}
	where ${\mathcal{N}}_T(\cdot)$ represents the {\em{truncated}} normal distribution whose nonzero probability density  concentrates in the positive orthant, ${\mathbf{P}} \triangleq \left[ {\begin{array}{*{20}{c}}
		\text{diag}({\mathbf{a}}) & -{\mathbf{W}} \\
		-{\mathbf{W}}^T & \text{diag}({\mathbf{d}})
		\end{array}} \right]\succ 0$ and ${\boldsymbol{\mu}} \triangleq {\mathbf{P}}^{-1} \left[ {\begin{array}{*{20}{c}}
		{\mathbf{b}} \\
		{\mathbf{c}} 
		\end{array}} \right]$. 
	Because of the diagonal matrices $\text{diag}({\mathbf{a}})$ and $\text{diag}({\mathbf{d}})$ in \eqref{energy_fun}, we have the conditional distributions as
	\begin{align}
		p({\mathbf{x}} | {\mathbf{h}}; {\boldsymbol{\Theta}} ) &= \prod_{i=1}^n {\mathcal{N}}_T\left(x_i \left| \frac{1}{a_i}[{\mathbf{W}}{\mathbf{h}}+{\mathbf{b}}]_i, \frac{1}{a_i} \right.\right), \label{p_x_h} \\
		p({\mathbf{h}} | {\mathbf{x}}; {\boldsymbol{\Theta}} ) &= \prod_{j=1}^m {\mathcal{N}}_T\left(h_j \left| \frac{1}{d_j}[{\mathbf{W}}^T{\mathbf{x}}+{\mathbf{c}}]_j, \frac{1}{d_j} \right.\right), \label{p_h_x}
	\end{align}
	where $[{\mathbf{z}}]_i$ and $z_i$ both represent the $i$-th element of vector ${\mathbf{z}}$. Equations (\ref{p_x_h}) and (\ref{p_h_x}) show that the visible variables are conditionally independent given the hidden variables, and vice versa.
	
	\begin{figure}
		\centering
		\includegraphics[width=0.3\textwidth]{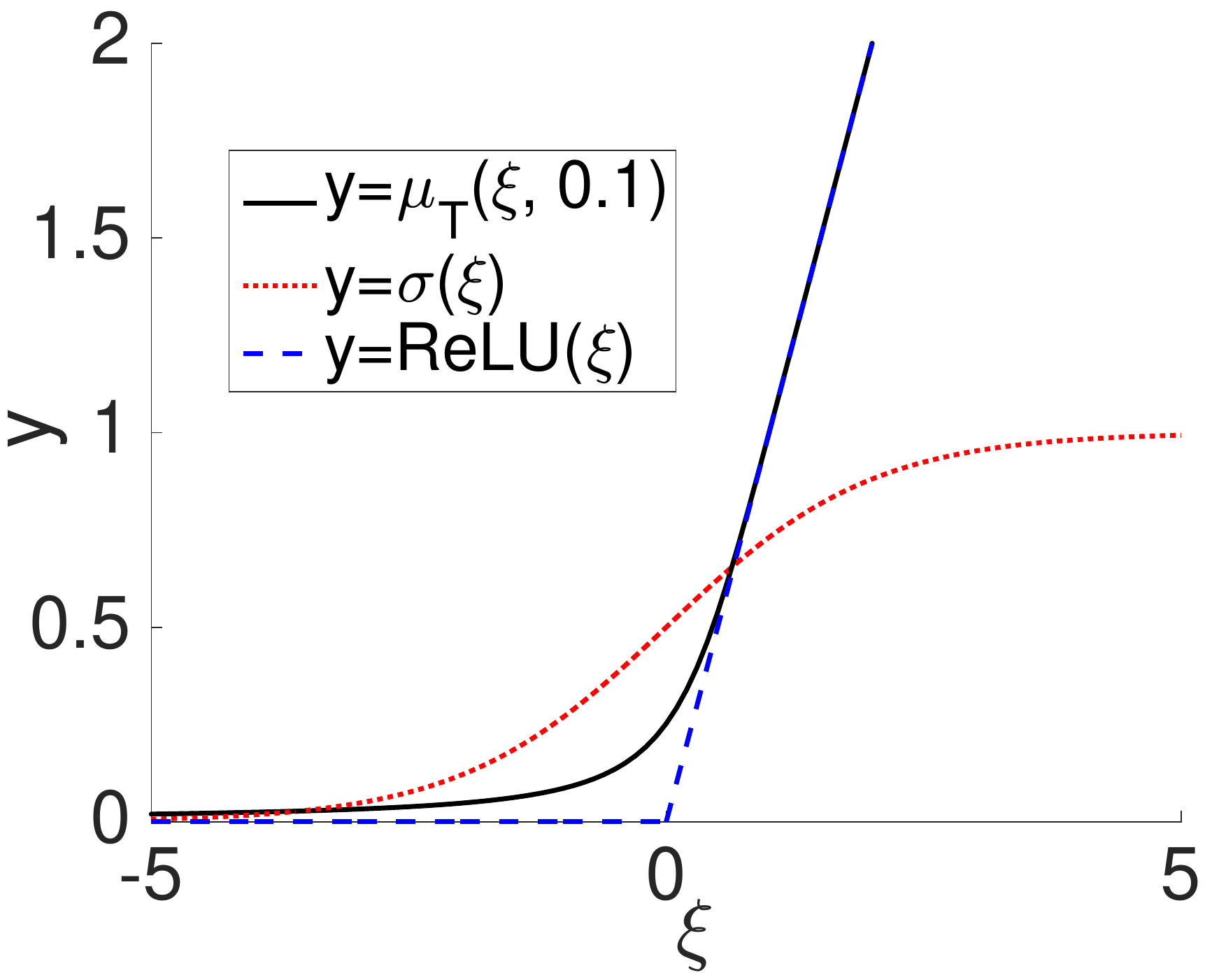}
		\caption{\small $\mu_T(\xi,0.1)$ vs $\sigma(\xi)=(1+e^{-\xi})^{-1}$ and $\mathrm{ReLU}(\xi)=\max(0,\xi)$.} \label{truncated_mean}
	\end{figure}

	By the properties of univariate truncated normal distributions \cite{johnson1994continuous}, the conditional expectation is given by ${\mathbb{E}}[h_j| {\mathbf{x}}] = \mu_T(\frac{1}{d_j}[{\mathbf{W}}^T{\mathbf{x}}+{\mathbf{c}}]_j, \frac{1}{{d_j}})$, where
	\begin{equation} \label{truncated_normal_mean}
		\mu_T(\xi, \lambda^2) \triangleq \xi + \lambda\,\phi\!\left(\xi/\lambda\right)/\,\Phi\!\left(\xi/\lambda\right)
	\end{equation}
	is the mean of ${\mathcal{N}}_T(x|\xi, \lambda^2)$ and it serves as the nonlinearity used in the RTGGM; $\phi(\cdot)$ and $\Phi(\cdot)$ are respectively the probability density function (pdf) and cumulative distribution function (cdf) of the standard normal distribution. Shown in Figure \ref{truncated_mean} is $\mu_T(\xi, \lambda^2)$ as a function of $\xi$ for $\lambda^2=0.1$, along with the sigmoidal and ReLU activation function, for comparison. It is observed that $\mu_T(\cdot,0.1)$ behaves similar to the ReLU nonlinearity, and deviates significantly from the sigmoidal nonlinearity used in RBMs.
	
	\subsection{Variants} \label{model_variant}
	Truncating the hidden variables in the RTGGM is essential to maintain model expressiveness. In the basic RTGGM above, the visible variables are also truncated to obtain symmetry, but this is not necessary. The visible domain can be changed to match the type of data in an application. Below we present three variants of the basic RTGGM, which deal with real, binary, and count data. In all cases, $p({\mathbf{x}}|{\mathbf{h}};{\boldsymbol{\Theta}})$ in (\ref{p_x_h}) is modified, but $p({\mathbf{h}}|{\mathbf{x}}, {\boldsymbol{\Theta}})$ remains as in \eqref{p_h_x}, and thus the ReLU nonlinearity is preserved.

	\paragraph{Real-Valued Data} The joint distribution in this case is $p({\mathbf{x}}, {\mathbf{h}}; {\boldsymbol{\Theta}}) = \frac{1}{Z} e^{- E({\mathbf{x}}, {\mathbf{h}})} {\mathbb{I}}({\mathbf{h}}\ge {\mathbf{0}})$. The conditional distribution of the data changes to
	\begin{equation} \label{p_x_h_real}
	p({\mathbf{x}}|{\mathbf{h}}; {\boldsymbol{\Theta}} ) = \prod_{i=1}^n {\mathcal{N}} \left(x_i \left| \frac{1}{a_i}[{\mathbf{W}}{\mathbf{h}}+{\mathbf{b}}]_i, \frac{1}{a_i} \right.\right).
	\end{equation}
	
	\paragraph{Binary Data}
	When each component of $\mathbf{x}$ is in $\{0,1\}$, the quadratic term ${\mathbf{x}}^T\text{diag}({\mathbf{a}}) {\mathbf{x}}$ is dropped from the energy function $E({\mathbf{x}}, {\mathbf{h}})$ and the domain restriction is changed from $\mathbb{I}(\mathbf{x}\geq\mathbf{0})$ to $\mathbb{I}(\mathbf{x} \in \{0, 1\}^n)$. The conditional in (\ref{p_x_h}) becomes $p(\mathbf{x}|{\mathbf{h}}; {\boldsymbol{\Theta}})=\prod_{i=1}^{n}p(x_i|{\mathbf{h}}; {\boldsymbol{\Theta}})$, with
	\begin{equation} \label{p_x_h_binary}
		p(x_i=1|{\mathbf{h}}; {\boldsymbol{\Theta}}) =  \frac{\exp\{[{\mathbf{W}}{\mathbf{h}}+{\mathbf{b}}]_i\}}{1+\exp\{[{\mathbf{W}}{\mathbf{h}}+{\mathbf{b}}]_i\}}.
	\end{equation}
	
	\paragraph{Count Data}
	Without loss of generality, we describe the count data model in the context of topic modeling. Following  \cite{hinton2009replicated}, we employ $N\times 1$ ``one-hot'' vectors (a one-hot vector is a vector of all $0$'s except for a single $1$) to represent the words in a vocabulary of size $N$. A document of size $K$ is then represented by a matrix ${\mathbf{X}}=[{\mathbf{x}}_1, \cdots, {\mathbf{x}}_K]$, where each column is a $N\times 1$ one-hot vector. We define an energy function $E({\mathbf{X}}, {\mathbf{h}}) \triangleq  \frac{1}{2}(\mathbf{h}^T\text{diag}({\mathbf{d}})\mathbf{h} - 2 \hat {\mathbf{x}}^T {\mathbf{W}}{\mathbf{h}} -2 {\mathbf{b}}^T\hat {\mathbf{x}} - 2K {\mathbf{c}}^T{\mathbf{h}})$ with $\hat {\mathbf{x}} \triangleq \sum_{k=1}^K {\mathbf{x}}_k$ understood as a count vector. The energy function above reduces to that of replicated softmax \cite{hinton2009replicated} if the quadratic term is dropped and ${\mathbf{h}}$ is restricted to $\mathbf{h}\in\{0,1\}^m$. The conditional in (\ref{p_x_h}) is accordingly modified to $p(\mathbf{X}|{\mathbf{h}}; {\boldsymbol{\Theta}})=\prod_{i=1}^{N}\prod_{k=1}^{K}p([{\mathbf{x}}_k]_i|{\mathbf{h}}; {\boldsymbol{\Theta}})$, with 
	\begin{eqnarray}\label{p_x_h_count}
		p([{\mathbf{x}}_k]_i = 1| {\mathbf{h}}; {\boldsymbol{\Theta}}) = \frac{\exp\{[{\mathbf{W}}{\mathbf{h}}+{\mathbf{b}}]_i \}}{\sum_{j=1}^N \exp\{[{\mathbf{W}}{\mathbf{h}}+{\mathbf{b}}]_j \}}.
	\end{eqnarray}

	\section{Model Training}
	\label{training-section}
	When training an RTGGM one is concerned with finding the $\boldsymbol{\Theta}$ that maximizes the log-likelihood $\widetilde{\mathcal{L}}(\boldsymbol{\Theta};\mathcal{X})=\sum_{\mathbf{x}\in\mathcal{X}}{\mathcal{L}}(\boldsymbol{\Theta};\mathbf{x}) $ given the training data set $\mathcal{X}$, where ${\mathcal{L}}(\boldsymbol{\Theta};\mathbf{x}) =\log \int_{0}^{+\infty}{p({\mathbf{x}}, {\mathbf{h}}; {\boldsymbol{\Theta}}) d{\mathbf{h}} }$ is the contribution from a single data sample, and $\int_{0}^{+\infty}d{\mathbf{h}}$ is a shorthand for the multiple integral with respect to (w.r.t.) the components in $\mathbf{h}$. It is known that $\frac{\partial {\mathcal{L}}(\boldsymbol{\Theta};\mathbf{x})}{\partial {\boldsymbol{\Theta}}} = {\mathbb{E}}[\frac{\partial E({\mathbf{x}}, {\mathbf{h}})}{\partial {\boldsymbol{\Theta}}}] - {\mathbb{E}}[\frac{\partial E({\mathbf{x}}, {\mathbf{h}})}{\partial {\boldsymbol{\Theta}}} | {\mathbf{x}}]$. The first term involves expectation  w.r.t. the model distribution $p(\mathbf{x},\mathbf{h})$, which is difficult due to the high variance inherent in the model distribution. Fortunately, we can resort to contrastive divergence (CD) to estimate the gradient. Specifically, starting with ${\mathbf{x}}^{(0)}={\mathbf{x}}$, the Gibbs sampler generates a chain of samples, $(\mathbf{x}^{(0)},\mathbf{h}^{(1)},\mathbf{x}^{(1)},\ldots,\mathbf{h}^{(k)},\mathbf{x}^{(k)})$, where $\mathbf{h}^{(t)}\sim{}p({\mathbf{h}}|{\mathbf{x}^{(t-1)}}; {\boldsymbol{\Theta}})$ and $\mathbf{x}^{(t)}\sim{}p({\mathbf{x}}|{\mathbf{h}^{(t)}}; {\boldsymbol{\Theta}})$. The contrastive divergence uses the first and last sample of $\mathbf{x}$ in the chain, i.e., $\mathbf{x}^{(0)}$ (which is a datum) and $\mathbf{x}^{(k)}$, to from an estimate of the expected gradient, 
	\begin{equation} \label{grad_CD}
		\frac{\partial {\mathcal{L}}(\boldsymbol{\Theta};\mathbf{x})}{\partial {\boldsymbol{\Theta}}} \approx {\mathbb{E}}\left[\frac{\partial E({\mathbf{x}}, {\mathbf{h}})}{\partial {\boldsymbol{\Theta}}}\left| {\mathbf{x}}^{(k)} \right.\right] - {\mathbb{E}}\left[\left.\frac{\partial E({\mathbf{x}}, {\mathbf{h}})}{\partial {\boldsymbol{\Theta}}} \right| {\mathbf{x}}^{(0)} \right].
	\end{equation}
	Note that $p({\mathbf{h}}|{\mathbf{x}}; {\boldsymbol{\Theta}})$ is always a truncated normal distribution as shown in \eqref{p_h_x}, while $p({\mathbf{x}}| {\mathbf{h}}; {\boldsymbol{\Theta}})$ is constituted according to \eqref{p_x_h}, \eqref{p_x_h_real}, \eqref{p_x_h_binary}, or \eqref{p_x_h_count}, depending on the type of data $\mathbf{x}$.  For the basic RTGGM, we  have $\frac{\partial E({\mathbf{x}}, {\mathbf{h}})}{\partial w_{ij}} = x_i h_j$, $\frac{\partial E({\mathbf{x}}, {\mathbf{h}})}{\partial a_i} = \frac{1}{2} x_i^2$, $\frac{\partial E({\mathbf{x}}, {\mathbf{h}})}{\partial b_i} = x_i$,  $\frac{\partial E({\mathbf{x}}, {\mathbf{h}})}{\partial c_j} = h_j$, and $\frac{\partial E({\mathbf{x}}, {\mathbf{h}})}{\partial d_j} = \frac{1}{2} h_j^2$. It can be seen that, to estimate  $\frac{\partial {\mathcal{L}}(\boldsymbol{\Theta};\mathbf{x})}{\partial {\boldsymbol{\Theta}}}$, one only needs to know the conditional expectations ${\mathbb{E}}[h_i| {\mathbf{x}} = {\mathbf{x}}^{(s)}]$ and ${\mathbb{E}}[h_i^2| {\mathbf{x}}^{(s)}]$ for $s=0,k$. It follows from \eqref{truncated_normal_mean} that ${\mathbb{E}}\left[h_j| {\mathbf{x}}^{(s)}\right] = \mu_T(\frac{[{\mathbf{W}}^T{\mathbf{x}}^{(s)}+{\mathbf{c}}]_j}{d_j}, \frac{1}{d_j})$. To compute ${\mathbb{E}}\left[h_j^2| {\mathbf{x}}^{(s)} \right]$, we use the formula ${\mathbb{E}}\left[h_j^2| {\mathbf{x}}^{(s)} \right] = {\mathbb{E}}\left[h_j| {\mathbf{x}}^{(s)} \right]^2  + \text{Var} [h_j|{\mathbf{x}}^{(s)}]$, where 
	\begin{equation} \label{truncated_normal_var}
		\text{Var} \left[h_j|{\mathbf{x}}^{(s)} \right] = \frac{1}{d_j}\left(1- \beta_j \frac{\phi\left(\beta_j\right)}{\Phi\left(\beta_j\right)} - \frac{\phi^2\left(\beta_j\right)}{\Phi^2 \left(\beta_j\right)}\right)
	\end{equation}
	according to \cite{johnson1994continuous} with $\beta_i \triangleq \frac{[{\mathbf{W}}^T{\mathbf{x}}^{(s)} + {\mathbf{c}}]_j}{\sqrt{d_j}}$. The gradients for the variant RTGGM models can be estimated similarly. With these estimated gradients, the model parameters ${\boldsymbol{\Theta}}$ can be updated using stochastic optimization algorithms.  
	
	One challenge in training RTGGMs is how to efficiently sample from truncated normal distributions. Fortunately, because the variables in an RTGGM are conditionally independent, we only need to sample from univariate truncated normals, and such sampling has been investigated extensively. Many efficient algorithms have been proposed \cite{chopin2011fast,robert1995simulation}. Another challenge is how to efficiently calculate the ratio $\frac{\phi(\mu)}{\Phi(\mu)}$ in \eqref{truncated_normal_mean} and \eqref{truncated_normal_var}.  Direct calculation is expensive due to the integration involved in $\Phi(\mu)$. To compute it cheaply, we adopt the approach in \cite{su2016learning}, taking into consideration the approximations based on asymptotic expansions of the Gaussian hazard function.

	\section{Partition Function Estimation}
	\label{Evaluation}
	To evaluate model performance, we desire the partition function $Z$. By exploiting the bipartite structure in an RTGGM as well as the appealing properties of truncated normals, we use annealed importance sampling (AIS) \cite{salakhutdinov2008quantitative,neal2001annealed} to estimate $Z$. We here only focus on the RTGGM with binary data; details for the other data types are provided in the Supplementary Material. The joint distribution of the RTGGM for binary data can be represented as $p({\mathbf{x}}, {\mathbf{h}}; {\boldsymbol{\Theta}}) = \frac{1}{Z} e^{-\frac{1}{2}\left(\|{\mathbf{D}}^{\frac{1}{2}}{\mathbf{h}}\|^2 - 2{\mathbf{x}}^T{\mathbf{W}} {\mathbf{h}} - 2{\mathbf{b}}^T{\mathbf{x}} - 2{\mathbf{c}}^T{\mathbf{h}} \right)} I({\mathbf{x}}\in \{0, 1\}^n)I({\mathbf{h}}\ge {\mathbf{0}})$. After integrating out the hidden variable ${\mathbf{h}}$, we obtain $p({\mathbf{x}}; {\boldsymbol{\Theta}}) = \frac{1}{Z} p^*({\mathbf{x}}; {\boldsymbol{\Theta}})$, where

	\begin{equation}
	p^*({\mathbf{x}}; {\boldsymbol{\Theta}}) \triangleq e^{{\mathbf{b}}^T{\mathbf{x}}} \times \prod_{j=1}^m \frac{1}{\sqrt{d_j}} \frac{\Phi\left(\frac{[{\mathbf{W}}^T{\mathbf{x}}+{\mathbf{c}}]_j}{\sqrt{d_j}}\right)}{\phi\left(\frac{[{\mathbf{W}}^T{\mathbf{x}}+{\mathbf{c}}]_j}{\sqrt{d_j}}\right)}.
	\end{equation}
	Since $p^*({\mathbf{x}}; {\boldsymbol{\Theta}})$ is in closed-form, we only need calculate the partition function $Z$ to obtain  $p({\mathbf{x}}; {\boldsymbol{\Theta}})$. 
	
	Following the AIS procedure \cite{salakhutdinov2008quantitative,neal2001annealed}, we define two distributions $p_A({\mathbf{x}}, {\mathbf{h}}^A) = \frac{1}{Z_A} e^{-E_A({\mathbf{x}}, {\mathbf{h}}^A)}$ and $p_B({\mathbf{x}}, {\mathbf{h}}^B) = \frac{1}{Z_B} e^{-E_B({\mathbf{x}}, {\mathbf{h}}^B)}$, where $E_A({\mathbf{x}}, {\mathbf{h}}^A) \triangleq \frac{1}{2}(\|\text{diag}^{\frac{1}{2}}({\mathbf{d}}){\mathbf{h}}^A\|^2 - 2 {\mathbf{b}}^{T}{\mathbf{x}})$ and  $E_B({\mathbf{x}}, {\mathbf{h}}^B) \triangleq \frac{1}{2}(\|\text{diag}^{\frac{1}{2}}({\mathbf{d}}){\mathbf{h}^B}\|^2 - 2{\mathbf{x}}^T{\mathbf{W}} {\mathbf{h}^B} - 2{\mathbf{b}}^T{\mathbf{x}} - 2{\mathbf{c}}^T{\mathbf{h}^B})$.  By construction, $p_0({\mathbf{x}}, {\mathbf{h}}^A, {\mathbf{h}}^B)=p_A({\mathbf{x}}, {\mathbf{h}}^A)$ and $p_K({\mathbf{x}}, {\mathbf{h}}^A, {\mathbf{h}}^B)=p_B({\mathbf{x}}, {\mathbf{h}}^B)$. The partition function of $p_A({\mathbf{x}}, {\mathbf{h}}^A)$ is given by $Z_A = \prod_{i=1}^n(1+e^{b_i^A}) \prod_{j=1}^m \frac{1}{\sqrt{d_j}} \frac{\Phi(0)}{\phi(0)}$, and the partition function of $p_B({\mathbf{x}}, {\mathbf{h}}^B)$ can be approximated as \cite{neal2001annealed}
	\begin{equation}
	Z_B \approx \frac{\sum_{i=1}^M w^{(i)}}{M}\,Z_A,
	\end{equation}
	where $w^{(i)}$ is constructed from a Markov chain that gradually transits from $p_A({\mathbf{x}}, {\mathbf{h}}^A)$ to $p_B({\mathbf{x}}, {\mathbf{h}}^B)$, with the transition realized via a sequence of intermediate distributions,
	\begin{equation}
	p_k({\mathbf{x}}, {\mathbf{h}}^A, {\mathbf{h}}^B) \!=\! \frac{1}{Z_k}e^{ - (1-\beta_k) E_A({\mathbf{x}}, {\mathbf{h}}^A) - \beta_k E_B({\mathbf{x}}, {\mathbf{h}}^B)},
	\end{equation}
	where $0=\beta_0<\beta_1<\ldots<\beta_{K}=1$. In particular, the Markov chain $(\mathbf{x}_i^{(0)},\mathbf{x}_i^{(1)},\ldots,\mathbf{x}_i^{(K)})$ is simulated as $\mathbf{x}_i^{(0)}\sim{}p_0({\mathbf{x}_i}, {\mathbf{h}}^A, {\mathbf{h}}^B)$, $({\mathbf{h}}^A, {\mathbf{h}}^B)\sim{}p_1({\mathbf{h}}^A, {\mathbf{h}}^B| {\mathbf{x}_i}^{(0)})$, $\mathbf{x}_i^{(1)}\sim{}p_1({\mathbf{x}_i}| {\mathbf{h}}^A, {\mathbf{h}}^B)$, $\cdots$, $({\mathbf{h}}^A, {\mathbf{h}}^B)\sim{}p_K({\mathbf{h}}^A, {\mathbf{h}}^B| {\mathbf{x}_i}^{(K-1)})$ and $\mathbf{x}_i^{(K)}\sim{}p_K({\mathbf{x}_i}| {\mathbf{h}}^A, {\mathbf{h}}^B)$.  From  the chain, a coefficient is constructed as $w^{(i)} = \frac{p_1^*(\tilde {\mathbf{x}_i}^{(0)}}{p_0^*(\tilde {\mathbf{x}_i}^{(0)})} \frac{p_2^*(\tilde {\mathbf{x}_i}^{(1)})}{p_1^*(\tilde {\mathbf{x}_i}^{(1)})} \cdots \frac{p_K^*(\tilde {\mathbf{x}_i}^{(K-1)})}{p_{K-1}^*(\tilde{\mathbf{x}_i}^{(K-1)})}$, where 
	\begin{align}
	p^*_k({\mathbf{x}}) & =  e^{(1-\beta_k){\mathbf{b}}^{AT}{\mathbf{x}}} \prod_{j=1}^m \frac{1}{\sqrt{(1-\beta_k)d_j}} \frac{\Phi\left(0\right)}{\phi\left(0\right)}  \nonumber \\
	&\quad \times e^{\beta_k{\mathbf{b}}^{T}{\mathbf{x}}} \prod_{j=1}^m \frac{1}{\sqrt{\beta_k d_j}} \frac{\Phi\left(\frac{\sqrt{\beta_k}[{\mathbf{W}}^{T}{\mathbf{x}} + {\mathbf{c}}]_j}{\sqrt{d_j }}\right)}{\phi\left(\frac{\sqrt{\beta_k}[{\mathbf{W}}^{T}{\mathbf{x}} + {\mathbf{c}}]_j}{\sqrt{d_j}}\right)}.
	\end{align}
	Assuming $M$ independent Markov chains simulated in this way, one obtains $\{w^{(i)}\}_{i=1}^M$. Note that the Markov chains  can be efficiently simulated, as all involved variables are conditionally independent. 
	
	\section{Extension to Deep Models}\label{Deep_model}
		\begin{figure}[t]
			\centering
			\includegraphics[width=0.43\textwidth]{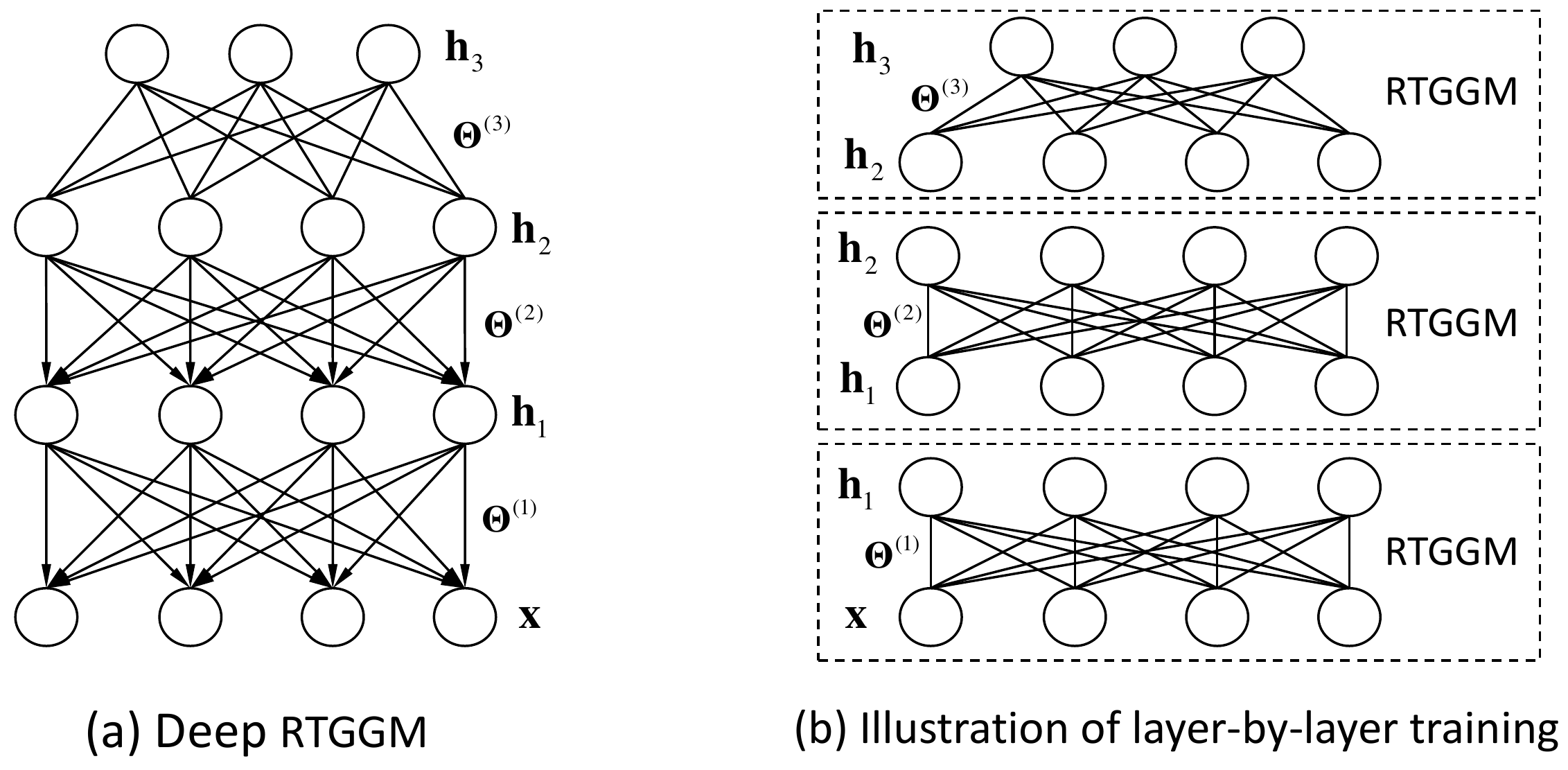}
			\caption{ \small (a) A deep RTGGM with three hidden layers. (b) Layer-wise training of three two-layer RTGGMs.} \label{Fig_DeepRTGGM}
			\vspace{-4mm}
		\end{figure}

	The RTGGMs discussed so far consist of a visible layer and a hidden layer. These two-layer models, like RBMs, can be used to construct deep models. A deep RTGGM with $L$ hidden layers, constructed by stacking $L$ two-layer RTGGMs, can be defined by the following joint distribution,
	\begin{align}
		p({\mathbf{h}}_L, \!\cdots, \! {\mathbf{h}}_1, {\mathbf{x}}) \!=\! p({\mathbf{h}}_L, \! {\mathbf{h}}_{L\!-\!1})  \cdots p({\mathbf{h}}_1|{\mathbf{h}}_2) p({\mathbf{x}}| {\mathbf{h}}_1),
	\end{align}
	where $p({\mathbf{h}}_L, {\mathbf{h}}_{L-1})$ is the joint distribution of a two-layer RTGGM 
	and $p({\mathbf{h}}_{\ell-1}| {\mathbf{h}}_{\ell}) = \prod_{i=1}^{M_{\ell-1}}{\mathcal{N}}_T( [{\mathbf{h}}_{\ell-1}]_i | \frac{1}{a^{(\ell)}_i}[{\mathbf{W}}^{(\ell)}{\mathbf{h}}_\ell + {\mathbf{b}}^{(\ell)} ]_i, \frac{1}{a_i^{(\ell)}} )$ is the associated conditional distribution. The bottom layer  $p({\mathbf{x}}| {\mathbf{h}}_1)$ could be defined by truncated normal, normal, binary or count distributions, depending on the data type. A deep RTGGM with three hidden layers is illustrated in the left panel of  Figure \ref{Fig_DeepRTGGM}.
	
	Similar to a DBN constructed from RBMs \cite{hinton2006fast}, a deep RTGGM can be trained in a layer-wise fashion. Specifically, we first train the bottom layer by simply treating it as an RTGGM, using the CD-based ML algorithm described above. We then compute the conditional expectation ${\mathbb{E}}[{\mathbf{h}}_1|{\mathbf{x}}]$ from the already-trained bottom RTGGM and use ${\mathbb{E}}[{\mathbf{h}}_1|{\mathbf{x}}]$ as data to train the second layer from the bottom, again treating it as a RTGGM. The layer-wise training procedure proceeds until the top layer is reached, as illustrated in the right panel of Figure \ref{Fig_DeepRTGGM}. Similar to the proofs in \cite{hinton2006fast}, we can prove that the variational lower bound is guaranteed to increase as more layers are added under the layer-wise training.
	
	Besides serving as a generative model, the deep RTGGM can also be used to pretrain a feedforward neural network so as to improve its performance. It is known that, due to the sigmoidal nonlinearity inherent in RBMs, when we use the unsupervised learning result of a DBN to initialize sigmoidal feed-forward neural networks, remarkable improvements are observed, especially in the case of scarce labeled data \cite{hinton2006reducing}. Due to the similarity between the nonlinearity in RTGGMs and $ReLU(\cdot)$, we can also use the deep RTGGM learned unsupervisedly to initialize ReLU feedforward neural networks.

	\section{Experiments}
	We report experimental results of the RTGGM models on various publicly available data sets, including binary, count and real-valued data, and compare them to competing models. For all RTGGM models considered below, we use $\mathbf{x}^{(0)}$ and $\mathbf{x}^{(25)}$ to get a CD-based gradient estimate and then use RMSprop to update the model parameters, with the RMSprop delay set to $0.95$.

	\begin{figure*}[t]
		\centering
		\begin{subfigure}
			\centering
			\includegraphics[width=0.23\textwidth]{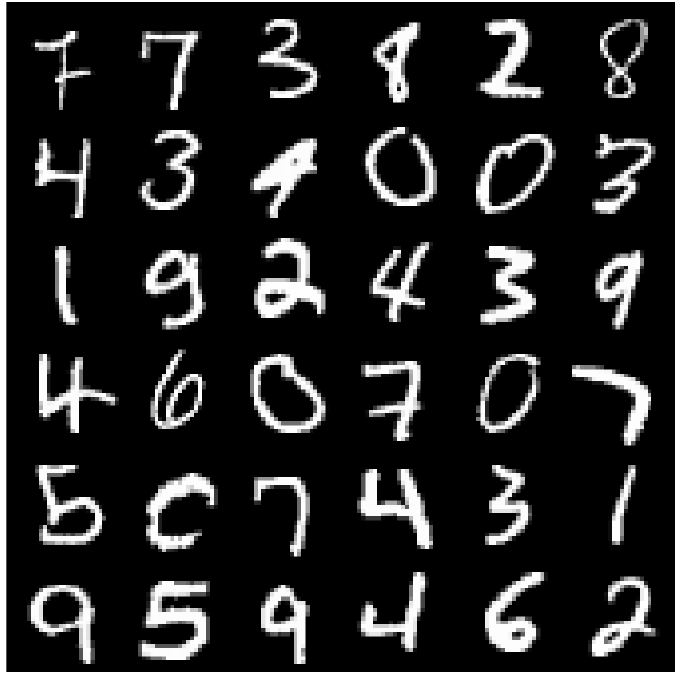}
		\end{subfigure}
		\begin{subfigure}
			\centering
			\includegraphics[width=0.23\textwidth]{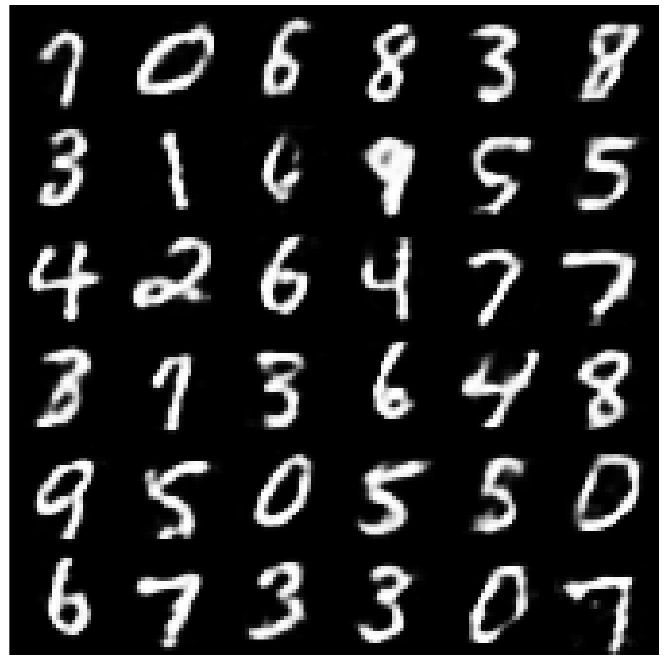}
		\end{subfigure}
		\begin{subfigure}
			\centering
			\includegraphics[width=0.34\textwidth]{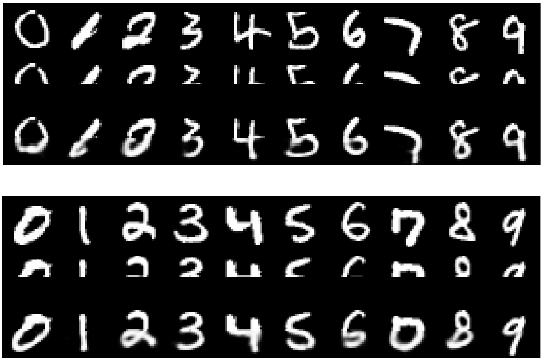}
		\end{subfigure}
		\caption{ \small (Left) Samples from the MNIST data set. (Middle) Samples drawn from the RTGGM with $500$ hidden nodes. (Right) In each sub-figure, the first row shows the samples from the testing data set, the second row shows the occluded digits presented to the RTGGM, and the bottom row shows the recovered digits. The grayscale values indicate the probabilities.} \label{Fig_MINIST_Gen}
	\end{figure*}

	\begin{figure*}[t]
		\centering
		\begin{subfigure}
			\centering
			\includegraphics[width=0.22\textwidth]{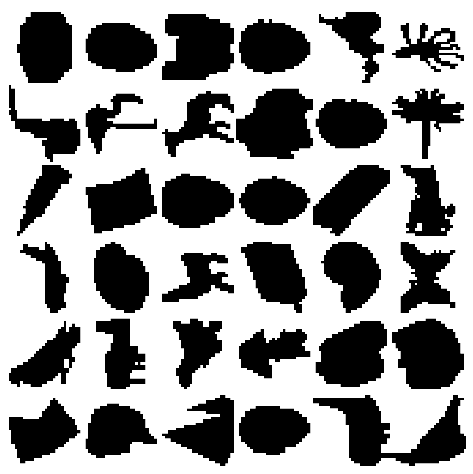}
		\end{subfigure}
		\begin{subfigure}
			\centering
			\includegraphics[width=0.22\textwidth]{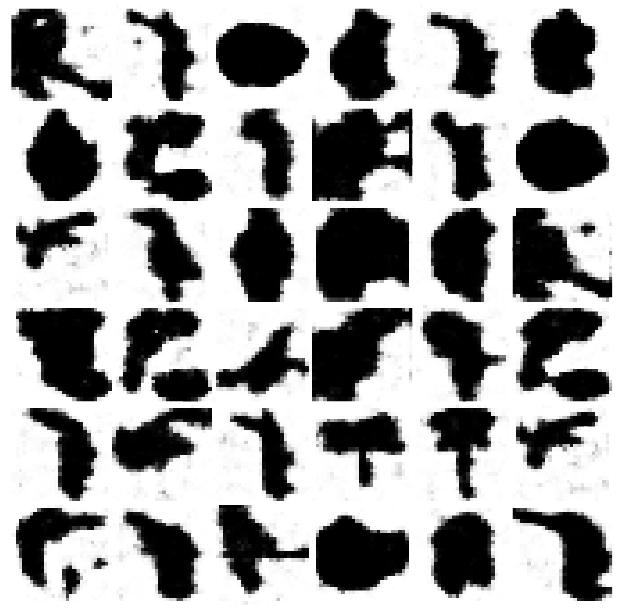}
		\end{subfigure}
		\begin{subfigure}
			\centering
			\includegraphics[width=0.28\textwidth]{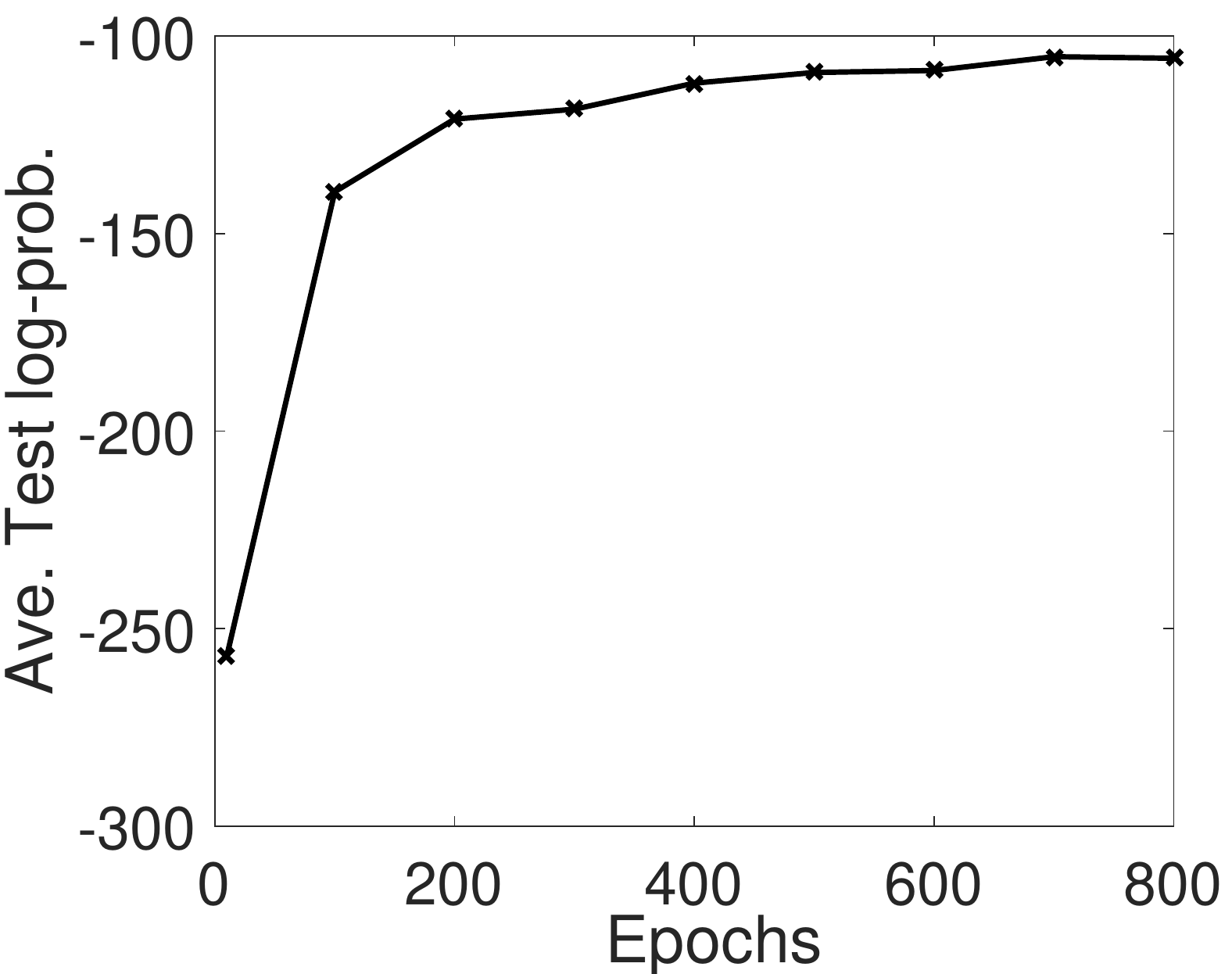}
		\end{subfigure}
		\caption{ \small (Left) Samples drawn from Caltech 101 Silhouettes data set. (Middle) Samples drawn from the RTGGM with $500$ hidden nodes. (Right) The average test log-probability of the RTGGM with 500 hidden nodes, as a function of learning epoch.} \label{Fig_Caltech_Gen}
	\end{figure*}

	\subsection{Binary Data}
	The binarized versions of MNIST and Caltech 101 Silhouettes data sets are considered. MNIST contains $28\times 28$ images of ten handwritten digits, with $60,\!000$ and $10,\!000$ images in the training and testing sets, respectively. Caltech 101 Silhouettes is a set of $28\times 28$ images for the polygon outlines of objects, of which $6364$ are used for training  and $2307$ for testing \cite{marlin2010inductive}. Two RTGGMs, with $100$ and $500$ hidden nodes, are trained and tested. The learning rate is set to $10^{-4}$ and the precision $d_i$ is set to $5$. Log-probabilities are estimated based on $100,\!000$ inverse temperatures $\beta_k$, uniformly spaced in $[0, 1]$. The final estimate is an average over $100$ independent AIS runs.
	
	Tables \ref{MNIST-table} and \ref{Caltech-table} summarize the average test log-probabilities on MNIST and Caltech 101 Silhouettes. For comparison, the corresponding results of competing models are also presented. It is seen from Table \ref{MNIST-table} that  the RTGGM with 500 hidden nodes achieves the best performance, significantly outperforming the RBM with the same number of hidden nodes as well as the deep SBN and DBN models. Similar results are observed in Table \ref{Caltech-table} for the Caltech 101 Silhouettes. The performance gain may be largely attributed to the smooth ReLU nonlinearity brought by truncation, as well as less approximations made in training. The importance of truncation is also revealed in the results of restricted GGMs (RGGM), in which we do not truncate the hidden variables but only impose bipartite structure on GGMs. It can be seen from both Tables \ref{MNIST-table} and \ref{Caltech-table} that, without the truncation, RGGMs perform poorly compared to all models. Note that the models in \cite{nair2010rectified,ravanbakhsh2016stochastic} are not included here, because their log-probability cannot be computed, as explained earlier. 
	
	To demonstrate that the RTGGM is able to capture important statistical relations, we show in Figure \ref{Fig_MINIST_Gen} samples drawn from the RTGGM trained on MNIST, using a Gibbs sampler with $50000$ burn-in samples. We see that the generated digits exhibit large variability and look very similar to real handwritten digits. Moreover, we also use the RTGGM to recover the missing values of an image. It is demonstrated in Figure \ref{Fig_MINIST_Gen} that, with only the upper-half part of an image presented, the RTGGM can recover the lower-half part reasonably well. The recovery is mostly correct except that ``2'' is mistaken for ``0'' in the upper subfigure and ``7'' is mistaken for ``0'' in the lower subfigure. However, we notice that the two cases are extremely difficult, in which it would be difficult even for a human to recognize the images based on only the upper-half parts. Finally, we demonstrate in Figure \ref{Fig_Caltech_Gen} the images drawn from the RTGGM trained on Caltech 101 Silhouettes, using Gibbs sampling with $50,\!000$ burn-in samples. Again, it can be seen that the generated images resemble the training data, showing that the generative model has faithfully captured the features of the training data.

	\begin{table} [!t]
			\centering
			\small
			\begin{tabular}{l l c}
				\toprule
				Model & Dim & Test log-prob. \\ \midrule
				RBM$^\star$ &  $500$ &  $-86.3$ \\
				SBN$^\circ$ &  $10$-$100$-$200$-$300$-$400$ & $-85.4$ \\
				DBN$^\diamond$ & $2000$-$500$ & $-86.2$ \\ 
				RGGM & 500& $-90.2$\\   \midrule
				RTGGM & $100$ &  $-89.3$\\
				RTGGM & $500$ &  $\boldsymbol{-83.2}$\\
				\bottomrule
			\end{tabular}
			\caption{\small Average test log-probability on MNIST.  ($\star$) results reported in \protect\cite{salakhutdinov2008quantitative} ; ($\circ$) results reported in  \protect\cite{bornschein2014reweighted}; ($\diamond$) results reported in  \protect\cite{hinton2006fast}.} \label{MNIST-table}
\end{table}

\begin{table}
			\centering
			\small
			\begin{tabular}{l l c}
				\toprule
				Model & Dim & Test log-prob. \\ \midrule
				RBM$^\star$ &  $500$ &  $-114.7$ \\
				RBM$^\star$ & $4000$ & $-107.7$ \\
				SBN$^\circ$ &  $10$-$50$-$100$-$300$ & $-113.3$ \\ 
				RGGM & 500& $-350.9$ \\   \midrule
				RTGGM & $100$ &  $-127.8$\\
				RTGGM & $500$ &  $\boldsymbol{-105.1}$\\
				\bottomrule
			\end{tabular}
		\caption{\small Average test log-probability on Caltech 101Silhouettes.  ($\star$) results reported in  \protect\cite{cho2013enhanced}; ($\circ$) results reported in  \protect\cite{bornschein2014reweighted}. }
		\label{Caltech-table}
	\end{table}

	\subsection{Count Data}
	Two publicly available corpora are considered: 20NewsGroups and Reuters Corpus Volume. The two corpora are preprocessed as in \cite{hinton2009replicated}. 
	An RTGGM with $50$ hidden nodes is trained, using the same setting as in the previous experiment for the learning rate. The perplexity is evaluated over 50 held-out documents, based on the setting used in \cite{hinton2009replicated}. For each document, we obtain the test log-probability as an average over 100 AIS runs, each using $100,\!000$ inverse temperatures $\beta_k$.
	
	Table \ref{count-table} shows the average test perplexity per word for the RTGGM. For comparison, we also report the perplexities of LDA with 50 and 200 topics, as well as those of the replicated softmax model (RSM) \cite{hinton2009replicated} with 50 topics. The RSM is a variant of the RBM that handles count data and it is constructed similarly as the RTGGM for count data. As seen from Table \ref{count-table}, for both corpora, RTGGM-50 performs better than RSM-50 and LDA models. This further demonstrates the performance gains brought about by the smooth ReLU nonlinearity in RTGGMs.
	
	\begin{table}[t]
		\small
		\centering
		\begin{tabular}{lccccc}
			\toprule
			Data set  
			& LDA-50$^{\star}$ & LDA-200$^{\star}$ & RSM-50$^{\star}$ & RTGGM-50 \\
			\midrule
			20news & $1091$ & $1058$ & $953$ &  ${\boldsymbol{915}}$\\
			Reuters & $1437$ & $1142$ & $988$ & ${\boldsymbol{934}}$\\
			\bottomrule
		\end{tabular}
				\caption{\small Average test perplexity per word, on 20newsgroup and Reuters. ($\star$) cited from \protect \cite{hinton2009replicated}. }
				\label{count-table}
\vspace{-2mm}
	\end{table}

	\subsection{Unsupervised Pre-training of ReLU Neural Networks} 	
	As described previously, deep RTGGMs can be used to pre-train multi-layer ReLU neural networks, by exploiting the unlabeled information. In this task, MNIST and NORB datasets are considered, where MNIST is the same as in previous experiments. NORB is a dataset of images from 6 classes on cluttered background, partitioned into  $291,\!600$ training and $58,\!230$ testing images. We pre-process these images as in \cite{nair2010rectified}. We first train a $1000$-$1000$-$1000$ deep RTGGM on the unlabeled data, and then use the learned model parameters to initialize a deep ReLU neural network of the same size, which is further trained with the provided labels using RMSprop, with a learning rate of $10^{-5}$. The test accuracy is reported in Table \ref{classification-table}. For comparison, we also report the results with no pre-training or pre-trained using the rectified-RBM \cite{nair2010rectified}. It is seen from Table \ref{classification-table} that the results of pre-training with the RTGGM performs the best. Note that this is an extension of \cite{hinton2006reducing}, where the RBM was used to pretrain a sigmoid-based neural network; here we use an RTGGM to pretrain a ReLU-based neural network.

	\begin{table}
		\centering
		\begin{tabular}{lccc}
			\toprule
			\multirow{2}{*}{Data set} & \multirow{2}{*}{Without pre-train} & \multicolumn{2}{c}{With pre-train}     \\
			\cmidrule{3-4}
			&  & rectified-RBM  &  RTGGM \\
			\midrule
			MNIST & $1.43\%$ &  $1.33\%$ & ${\mathbf{1.17\%}}$ \vspace{2.2mm}\\
			NORB & $16.88\%$ & $16.43\%$ & ${\mathbf{16.12\%}}$ \\
			\bottomrule
		\end{tabular}
\caption{\small Average classification errors achieved by the multi-layer ReLU neural network without pre-training, pre-trained by the method in (Nair and Hinton 2010) (referred to as ``rectified-RBM'' in the table), and pre-trained by the deep RTGGM. }	\label{classification-table}
	\end{table}

	\section{Conclusions}
	We have introduced a novel variant of the GGM, called restricted truncated GGM (RTGGM), to enhance its representational abilities while preserving its nice (simple inference) properties. The new model is obtained by truncating the variables of an undirected GGM and imposing a bipartite structure on the truncated GGM. It is shown that the truncation brings strong nonlinear representational power to the model, while the bipartite structure enables the model to be trained efficiently using contrastive divergence. Three variants of the RTGGM have been developed to handle real, binary and count data. The two-layer RTGGM has further been extended to produce deep models with multiple hidden layers. Methods have also been developed to estimate the partition function used in evaluating the unsupervised learning performance. Extensive experimental results have demonstrated the superior performance of RTGGMs in unsupervised learning of many types of data, as well as in unsupervised pre-training of feedforward ReLU neural networks.
\section*{Acknowledgements} The research reported here was supported by the DOE, NGA, NSF, ONR and by Accenture.

	\bibliographystyle{aaai} 
	\bibliography{reference}

	\newpage
	
	\appendix

	\twocolumn[
	\begin{center}
		\bf{\LARGE Supplementary Materials for ``Unsupervised Learning with Truncated Gaussian Graphical Models''\\ \vspace{7mm}}
	\end{center}
	]
	
	\section{Gradient Computation}
	
	From the expression of energy function, it can be seen that $\frac{\partial E({\mathbf{x}}, {\mathbf{h}})}{\partial w_{ij}} = x_i h_j$, $\frac{\partial E({\mathbf{x}}, {\mathbf{h}})}{\partial a_i} = \frac{1}{2} x_i^2$, $\frac{\partial E({\mathbf{x}}, {\mathbf{h}})}{\partial b_i} = x_i$,  $\frac{\partial E({\mathbf{x}}, {\mathbf{h}})}{\partial c_j} = h_j$ and $\frac{\partial E({\mathbf{x}}, {\mathbf{h}})}{\partial d_j} = \frac{1}{2} h_j^2$, where $w_{ij}$ is the $(i,j)$-th element of ${\mathbf{W}}$; and $x_i$, $h_i$, $a_i$, $b_i$, $c_i$ and $d_i$ are the $i$-th element of corresponding vectors. Then, the deritives of log-likelihood can easily be obtained as
	\begin{align}
		\frac{\partial {\mathcal{L}}({\boldsymbol{\Theta}}; {\mathbf{x}} )}{\partial w_{ij}} &= -\left({\mathbb{E}}\left[x_ih_j\right] - x_i {\mathbb{E}}\left[h_j|{\mathbf{x}}\right]\right), \\
		\frac{\partial {\mathcal{L}}({\boldsymbol{\Theta}}; {\mathbf{x}} )}{\partial a_{i}} &= \frac{1}{2}\left({\mathbb{E}}\left[x_i^2\right] - x_i^2\right), \\
		\frac{\partial {\mathcal{L}}({\boldsymbol{\Theta}}; {\mathbf{x}} )}{\partial b_i} &= -\left({\mathbb{E}}\left[x_i\right] - x_i \right), \\
		\frac{\partial {\mathcal{L}}({\boldsymbol{\Theta}}; {\mathbf{x}} )}{\partial c_j } &= -\left({\mathbb{E}}\left[h_j\right] - {\mathbb{E}}\left[h_j|{\mathbf{x}}\right] \right), \\
		\frac{\partial {\mathcal{L}}({\boldsymbol{\Theta}}; {\mathbf{x}} )}{\partial d_j } &= \frac{1}{2}\Big( {\mathbb{E}}\left[h_j^2 \right] - {\mathbb{E}}\left[h_j^2|{\mathbf{x}}\right]\Big),
	\end{align}
	where expectations ${\mathbb{E}}[\cdot]$ and ${\mathbb{E}}[\cdot| {\mathbf{x}}]$ are taken w.r.t. $p({\mathbf{x}}, {\mathbf{h}})$ and $p({\mathbf{h}}| {\mathbf{x}})$, respectively. Due to the difficulties of obtaining ${\mathbf{E}}[\cdot]$, we resort to contrastive divergence (CD) algorithm to estimate the gradients. To this end, we sample ${\mathbf{h}}$ given ${\mathbf{x}}$ from $p({\mathbf{h}}|{\mathbf{x}})$, and then sample ${\mathbf{x}}$ from the sampled ${\mathbf{\hat h}}$ according to $p({\mathbf{x}}|{\mathbf{\hat h}})$. We denote this sample as ${\mathbf{x}}^{(1)}$. By repeating this process $k$ steps, we obtain a sample denoted as ${\mathbf{x}}^{(k)}$. Then, we can estimate the derivatives as
	\begin{align}
		\frac{\partial {\mathcal{L}}({\boldsymbol{\Theta}}; {\mathbf{x}})}{\partial w_{ij}} &= \!-\!\left(x_i^{(k)} {\mathbb{E}}\left[h_j| {\mathbf{x}}^{(k)}\right] \!-\! x_i^{(0)} {\mathbb{E}}\left[h_j|{\mathbf{x}}^{(0)} \right] \right), \\
		\frac{\partial {\mathcal{L}}({\boldsymbol{\Theta}})}{\partial a_{i}} &= \frac{1}{2}\left( x_i^{(k)2} - x_i^{(0)2} \right), \\
		\frac{\partial {\mathcal{L}}({\boldsymbol{\Theta}}; {\mathbf{x}})}{\partial b_i} &= -\left(x_i^{(k)} - x_i^{(0)}\right), \\
		\frac{\partial {\mathcal{L}}({\boldsymbol{\Theta}}; {\mathbf{x}})}{\partial c_j } &= -\left({\mathbb{E}}\left[h_j|{\mathbf{x}}^{(k)} \right] - {\mathbb{E}}\left[h_j|{\mathbf{x}}^{(0)} \right]\right), \\
		\frac{\partial {\mathcal{L}}({\boldsymbol{\Theta}}; {\mathbf{x}})}{\partial d_j } &= \frac{1}{2}\Big( {\mathbb{E}}\left[h_j^2 | {\mathbf{x}}^{(k)} \right] - {\mathbb{E}}\left[h_j^2|{\mathbf{x}}^{(0)}\right]\Big),
	\end{align}
	where ${\mathbf{x}}^{(0)} \triangleq {\mathbf{x}}$.
	
	\section{Derivation of $p^*({\mathbf{x}}; {\boldsymbol{\Theta}})$ }
	From the joint distribution $p({\mathbf{x}}, {\mathbf{h}}; {\boldsymbol{\Theta}})$, we have
	\begin{align} \label{binary_p_star_pre}
		&p^*({\mathbf{x}}; {\boldsymbol{\Theta}}) \nonumber \\
		&\quad = \frac{1}{Z} e^{{\mathbf{b}}^T{\mathbf{x}}} \prod_{j=1}^m \int_{0}^{+\infty}\!\!\! e^{-\frac{1}{2}\left(d_jh_j^2 - 2\left[{\mathbf{W}}^T{\mathbf{x}} + {\mathbf{c}}\right]_j h_j\right)} d h_j \nonumber \\
		&\quad  = \frac{1}{Z} e^{{\mathbf{b}}^T{\mathbf{x}}} \prod_{j=1}^m e^{\frac{[{\mathbf{W}}^T{\mathbf{x}} + {\mathbf{c}}]_j^2}{2d_j}}  \!\! \int_0^{+\infty}{\!\!\! e^{-\frac{d_j}{2} \left(h_j - \frac{[{\mathbf{W}}^T{\mathbf{x}} + {\mathbf{c}}]_j}{d_j}\right)\!^2}\! dh_j} .
	\end{align}
	After some simple manipulations, we obtain
	\begin{align} \label{binary_p_star_final}
		\nonumber \\
		p^*({\mathbf{x}}; {\boldsymbol{\Theta}}) & = \frac{1}{Z} e^{{\mathbf{b}}^T{\mathbf{x}}} \prod_{j=1}^m e^{\frac{[{\mathbf{W}}^T{\mathbf{x}} + {\mathbf{c}}]_j^2 }{2d_j}} \! \left(\frac{2\pi}{d_j}\right)^{\frac{1}{2}} \! \Phi\left(\! \frac{[{\mathbf{W}}^T{\mathbf{x}} + {\mathbf{c}}]_j }{\sqrt{d_j}}\! \right) \nonumber \\
		& = \frac{1}{Z} e^{{\mathbf{b}}^T{\mathbf{x}}} \prod_{j=1}^m \frac{1}{\sqrt{d_j}} \frac{\Phi\left(\frac{[{\mathbf{W}}^T{\mathbf{x}} + {\mathbf{c}}]_j }{\sqrt{d_j}}\right)}{\phi\left(\frac{[{\mathbf{W}}^T{\mathbf{x}} + {\mathbf{c}}]_j }{\sqrt{d_j}}\right) },
	\end{align}
	where $\phi(\cdot)$ and $\Phi(\cdot)$ means the PDF and CDF of standard normmal distribution.

	\section{Partition Function Estimation with \\Count Data}
	Without loss of generality, the count model is described in the context of bag-of-words topic model.  Following  \cite{hinton2009replicated}, we use a matrix ${\mathbf{X}}$ to describe a document, in whch each column is a column hot-vector, with the non-zero element representing the corresponding word in vocabulary appearing once. Now, we have the joint pdf describing bag-of-words document ${\mathbf{X}}$ as
	\begin{align}
		p({\mathbf{X}}, {\mathbf{h}}; {\boldsymbol{\Theta}}) &= \frac{1}{Z} e^{-\frac{1}{2}\left(\|{\mathbf{D}}^{\frac{1}{2}}{\mathbf{h}}\|^2 - 2\hat {\mathbf{x}}^T{\mathbf{W}} {\mathbf{h}} - 2{\mathbf{b}}^T\hat {\mathbf{x}} - 2K {\mathbf{c}}^T{\mathbf{h}} \right)} \nonumber \\
		&\quad   I({\mathbf{x}}_k\in {\mathcal{I}} )I({\mathbf{h}}\ge {\mathbf{0}}),
	\end{align}
	where 
	\begin{equation}
		\hat {\mathbf{x}} \triangleq \sum_{k=1}^K {\mathbf{x}}_k
	\end{equation}
	with ${\mathbf{x}}_k$ being the $k$-th column of ${\mathbf{X}}$; and $K$ means the number of words in the document; ${\mathcal{I}}$ is the set of one-hot vectors. Similar to operations in \eqref{binary_p_star_pre} and \eqref{binary_p_star_final}, we obtain
	\begin{align}
		p({\mathbf{X}}; {\boldsymbol{\Theta}}) = \frac{1}{Z}e^{{\mathbf{b}}^T\hat {\mathbf{x}}} \times \prod_{j=1}^m \frac{1}{\sqrt{d_j}} \frac{\Phi\left(\frac{[{\mathbf{W}}^T\hat {\mathbf{x}}+K{\mathbf{c}}]_j}{\sqrt{d_j}}\right)}{\phi\left(\frac{[{\mathbf{W}}^T \hat {\mathbf{x}}+K{\mathbf{c}}]_j}{\sqrt{d_j}}\right)}.
	\end{align}

	To estimate the partition function $Z$, define $p_A({\mathbf{X}}, {\mathbf{h}}^A; {\boldsymbol{\Theta}}) = \frac{1}{Z_A} e^{-\frac{1}{2}||{\mathbf{D}}^{\frac{1}{2}}{\mathbf{h}}^A||^2} I({\mathbf{x}}_k\in {\mathcal{I}}) I({\mathbf{h}}^A \ge 0)$ and $p_B({\mathbf{X}}, {\mathbf{h}}^B; {\boldsymbol{\Theta}}) = p({\mathbf{X}}, {\mathbf{h}}^B; {\boldsymbol{\Theta}})$. It is known from \cite{neal2001annealed} that the partition function $Z$ can be estimated as
	\begin{equation}
		Z \approx \frac{\sum_{i=1}^M w^{(i)}}{M}\,Z_A;
	\end{equation}
	where $Z_A$ can be easily computed as
	\begin{equation}
		Z_A = \prod_{j=1}^m \frac{1}{\sqrt{d_j}} \frac{\Phi(0)}{\phi(0)},
	\end{equation}
	and coefficient $w^{(i)}$ is constructed from a Markov chain that is simulated to gradually transit from $p_A({\mathbf{x}}, {\mathbf{h}}^A)$ to $p_B({\mathbf{x}}, {\mathbf{h}}^B)$, with the transition realized via a sequence of intermediate distributions
	\begin{align}
		&p_s({\mathbf{X}}, {\mathbf{h}}^A, {\mathbf{h}}^B; {\boldsymbol{\Theta}}) \nonumber \\
		&\quad = \frac{1}{Z_s}\exp \Big\{-\frac{1}{2}(1-\beta_s)||{\mathbf{D}}^{\frac{1}{2}}{\mathbf{h}}^A||^2 -\frac{1}{2} \beta_s||{\mathbf{D}}^{\frac{1}{2}}{\mathbf{h}}^B||^2   \nonumber \\
		&\quad \quad +\beta_s\hat {\mathbf{x}}^T {\mathbf{W}}{\mathbf{h}}^B + \beta_s{\mathbf{b}}^T\hat{\mathbf{x}} + \beta_s K {\mathbf{c}}^T{\mathbf{h}}^B \Big\}\nonumber \\
		&\quad \quad \times I({\mathbf{x}}_k\in {\mathcal{I}} ) I({\mathbf{h}}^A, {\mathbf{h}}^B \ge {\mathbf{0}})
	\end{align}
	for $\beta_s\in [0, 1]$. Specifically, $w^{(i)}$ is computed as
	\begin{equation} \label{w_i}
		w^{(i)} = \frac{p_1^*( {\mathbf{x}_i}^{(0)}}{p_0^*( {\mathbf{x}_i}^{(0)})} \frac{p_2^*( {\mathbf{x}_i}^{(1)})}{p_1^*( {\mathbf{x}_i}^{(1)})} \cdots \frac{p_K^*( {\mathbf{x}_i}^{(K-1)})}{p_{K-1}^*({\mathbf{x}_i}^{(K-1)})},
	\end{equation}
	where $p_s^*({\mathbf{X}}; {\boldsymbol{\Theta}})$ is the unnormalized term in $p_s({\mathbf{X}}; {\boldsymbol{\Theta}}) = \frac{1}{Z_s} p_s^*({\mathbf{X}}; {\boldsymbol{\Theta}}) I({\mathbf{x}}_k\in {\mathcal{I}} )$. To compute $w^{(i)}$, we integrate out ${\mathbf{h}}^A$ and ${\mathbf{h}}^B$ in $p_s({\mathbf{X}}, {\mathbf{h}}^A, {\mathbf{h}}^B; {\boldsymbol{\Theta}})$ and obtain
	\begin{align}
		\log p_s^*({\mathbf{X}}; {\boldsymbol{\Theta}}) & = - \frac{1}{2}\sum_{j=1}^m \log((1-\beta_s)d_j ) - m\log \frac{\phi(0)}{\Phi(0)} \nonumber \\
		&\quad + \beta_s{\mathbf{b}}^T \hat {\mathbf{x}} -\frac{1}{2}\sum_{j=1}^m \log (\beta_s d_j) \nonumber \\
		&\quad - \sum_{j=1}^m \log \frac{\phi\left(\frac{\sqrt{\beta_s}[{\mathbf{W}}^{T} \hat {\mathbf{x}} + K{\mathbf{c}} ]_j}{\sqrt{d_j }}\right)}{\Phi\left(\frac{\sqrt{\beta_s}[{\mathbf{W}}^{T} \hat {\mathbf{x}} + K {\mathbf{c}}]_j}{\sqrt{d_j}}\right)}.
	\end{align}
	In \eqref{w_i}, the sequence ${\mathbf{x}}^{(k)}$ is simulated from a Markov chain $(\mathbf{x}_i^{(0)},\mathbf{x}_i^{(1)},\ldots,\mathbf{x}_i^{(K)})$ as $\mathbf{x}_i^{(0)}\sim{}p_0({\mathbf{x}_i}, {\mathbf{h}}^A, {\mathbf{h}}^B)$, $({\mathbf{h}}^A, {\mathbf{h}}^B)\sim{}p_1({\mathbf{h}}^A, {\mathbf{h}}^B| {\mathbf{x}_i}^{(0)})$, $\mathbf{x}_i^{(1)}\sim{}p_1({\mathbf{x}_i}| {\mathbf{h}}^A, {\mathbf{h}}^B)$, $\cdots$, $({\mathbf{h}}^A, {\mathbf{h}}^B)\sim{}p_K({\mathbf{h}}^A, {\mathbf{h}}^B| {\mathbf{x}_i}^{(K-1)})$ and $\mathbf{x}_i^{(K)}\sim{}p_K({\mathbf{x}_i}| {\mathbf{h}}^A, {\mathbf{h}}^B)$, with the conditional pdfs equal to
	\begin{align}
		p({\mathbf{h}}^B|  {\mathbf{x}}; {\boldsymbol{\Theta}} ) & = {\mathcal{N}}_T\left({\mathbf{h}}^B \left| {\mathbf{D}}^{-1}({\mathbf{W}}^{T}{\mathbf{x}} + K{\mathbf{c}}), \frac{1}{\beta_s}{\mathbf{D}}^{-1} \right. \right), \\
		p( {\mathbf{x}}| {\mathbf{h}}^B; {\boldsymbol{\Theta}}) &= Multinomial\left( {\mathbf{x}}; K, {\boldsymbol{\xi}} \right),
	\end{align}
	where $\xi_i \triangleq \frac{\exp\{\beta_s [{\mathbf{W}}{\mathbf{h}}^B +{\mathbf{b}}]_i \}}{\sum_{i=1}^N \exp\{\beta_s [{\mathbf{W}}{\mathbf{h}}^B +{\mathbf{b}}]_i \}}$ is the success probability of the $i$-th word; and $K$ is the number of words in the document. Moreover, we can also derive that

	\section{Missing Data Prediction}
	Suppose the data ${\mathbf{x}}$ is composed of the observed part ${\mathbf{x}}_o$ and the unobserved part ${\mathbf{x}}_u$. With the help of already trained RTGGM model, we will recover the missing part ${\mathbf{x}}_u$ from the observed data ${\mathbf{x}}_o$. Obviously, the conditional pdf $p({\mathbf{x}}_u, {\mathbf{h}}| {\mathbf{x}}_o)$ constitutes a new RTGGM, with its energy function expressed as
	\begin{align}
		E({\mathbf{x}}_u, {\mathbf{h}}) &= \frac{1}{2}\Big(||{\mathbf{A}}^{\frac{1}{2}}_u {\mathbf{x}}_u ||^2 + ||{\mathbf{D}}^{\frac{1}{2}}{\mathbf{h}}||^2 - 2{\mathbf{x}}_u^T{\mathbf{W}}_u{\mathbf{h}} \nonumber \\
		&\quad - 2{\mathbf{b}}_u^T{\mathbf{x}}_u - 2{\mathbf{\tilde c}}^T {\mathbf{h}} \Big)
	\end{align}
	where ${\mathbf{\tilde c}} \triangleq {\mathbf{c}} + {\mathbf{W}}_o^T{\mathbf{x}}_o$ and ${\mathbf{A}}_u \triangleq \text{diag}({\mathbf{a}}_u)$; and ${\mathbf{a}}_u$, ${\mathbf{W}}_u$ and ${\mathbf{b}}_u$ are composed of partial rows of ${\mathbf{a}}$, ${\mathbf{W}}$ and ${\mathbf{b}}_u$, respectively, with the selected row indexes corresponding to those of ${\mathbf{x}}_u$. Hence, under the Gaussian output case, we have the pdf $p({\mathbf{x}}_u, {\mathbf{h}}| {\mathbf{x}}_o)$ as
	\begin{equation}
		p({\mathbf{x}}_u, {\mathbf{h}}| {\mathbf{x}}_o) \propto e^{-E({\mathbf{x}}_u, {\mathbf{h}})} {\mathbb{I}}({\mathbf{h}}\ge {\mathbf{0}}).
	\end{equation}
	The conditional pdfs can be further derived as
	\begin{align}
		p({\mathbf{x}}_u|{\mathbf{h}}, {\mathbf{x}}_o) &= {\mathcal{N}}\left({\mathbf{x}}_u| {\mathbf{A}}_u^{-1}({\mathbf{W}}_u{\mathbf{h}}+{\mathbf{b}}_u ), {\mathbf{A}}_u^{-1} \right), \\
		p({\mathbf{h}}|{\mathbf{x}}_u, {\mathbf{x}}_o) &= {\mathcal{N}}_T\left({\mathbf{h}}| {\mathbf{D}}^{-1}({\mathbf{W}}_u^T{\mathbf{x}}_u + {\mathbf{\tilde c}}), {\mathbf{D}}^{-1} \right).
	\end{align}
	By resorting to Gibbs sampling, the expectation ${\mathbb{E}}[{\mathbf{x}}_u|{\mathbf{x}}_o]$ can be estimated efficiently thanks to the conditional independence among ${\mathbf{x}}_u$ and ${\mathbf{h}}$. Notice that here we only considered the model with Gaussian observation, but it can be easily extended to binary observation models.

	\section{Restricted Gaussian Graphical Model}
	When we restrict the GGM having a bipartite structure and binary output, the joint pdf can be represented as
	\begin{equation}
		p({\mathbf{x}}, {\mathbf{h}}; {\boldsymbol{\Theta}}) \propto e^{- E({\mathbf{x}}, {\mathbf{h}})} {\mathbb{I}}\left({\mathbf{x}}\in \{0, 1\}^N\right),
	\end{equation}
	where
	\begin{align}
		E({\mathbf{x}}, {\mathbf{h}}) \triangleq \frac{1}{2}\left(  {\mathbf{h}}^T\text{diag}({\mathbf{d}}){\mathbf{h}} \!-\! 2{\mathbf{x}}^T{\mathbf{W}} {\mathbf{h}} \!-\! 2{\mathbf{b}}^T{\mathbf{x}} \!-\! 2{\mathbf{c}}^T{\mathbf{h}} \right).
	\end{align}
	The model can be trained efficiently using contrastive divergence (CD), with the conditional pdfs given by
	\begin{align}
		p({\mathbf{h}}|{\mathbf{x}}) & = {\mathcal{N}}\left(\text{diag}^{-1}({\mathbf{d}})\left({\mathbf{W}}^T{\mathbf{x}} \!+\! {\mathbf{c}}\right), \text{diag}^{-1}({\mathbf{d}})\right),  \\
		p({\mathbf{x}}| {\mathbf{h}}) &= \prod_{i=1}^N \frac{\left[{\mathbf{W}}{\mathbf{h}}+ {\mathbf{b}}\right]_i x_i}{1+\exp\{\left[{\mathbf{W}}{\mathbf{h}}+ {\mathbf{b}}\right]_i\}}.
	\end{align}
	After training the model, we use AIS to evaluate its performance. The derivation process is almost the same as that in RTGGM models except the marginal pdf has the form
	\begin{align}
		p({\mathbf{x}}; {\boldsymbol{\Theta}}) &= \frac{1}{Z} e^{{\mathbf{b}}^T{\mathbf{x}} }  \prod_{j=1}^m \int_{-\infty}^{+\infty}e^{-\frac{1}{2}\left(d_jh_j^2 - 2\left[{\mathbf{W}}^T{\mathbf{x}} + {\mathbf{c}}\right]_j h_j\right)} d h_j \nonumber \\
		& = \frac{1}{Z} e^{{\mathbf{b}}^T{\mathbf{x}}}  \prod_{j=1}^m e^{\frac{[{\mathbf{W}}^T{\mathbf{x}} + {\mathbf{c}}]_j^2}{2d_j}} \!\!  \int_{-\infty}^{+\infty}{\!\!\! e^{-\frac{d_j}{2} \left(h_j - \frac{[{\mathbf{W}}^T{\mathbf{x}} + {\mathbf{c}}]_j}{d_j}\right)^2} \!\! dh_j}  \nonumber \\
		& = \frac{1}{Z} e^{{\mathbf{b}}^T{\mathbf{x}}}  \prod_{j=1}^m e^{\frac{[{\mathbf{W}}^T{\mathbf{x}} + {\mathbf{c}}]_j^2 }{2d_j}} \left(\frac{2\pi}{d_j}\right)^{\frac{1}{2}} \nonumber \\
		& = \frac{1}{Z} e^{{\mathbf{b}}^T{\mathbf{x}}}  \prod_{j=1}^m \frac{1}{\sqrt{d_j} \phi\left(\frac{[{\mathbf{W}}^T{\mathbf{x}} + {\mathbf{c}}]_j }{\sqrt{d_j}}\right) }.
	\end{align}

	\section{Generation with Two-layer RTGGM}
	
\begin{figure}[h]
		\includegraphics[width=0.45\textwidth]{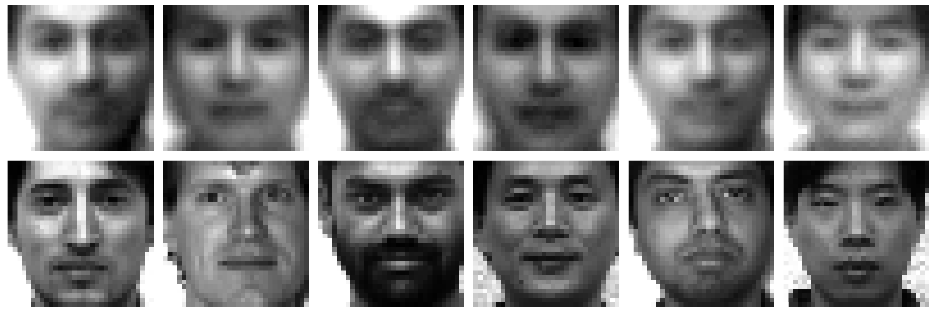}		
		\caption{(Top) Faces drawn from the deep RTGGM with two-layer hidden structure $10$-$100$. (Bottom) The corresponding most similar faces in the Yale Face data set.} \label{face_fig}					
	\end{figure}

	Yale Face data set is considered, which contains faces of $15$ persons, with each person having $11$  images of size $32\times 32$ under different expressions and lighting. Two-hidden-layer RTGGM ($10$-$100$) is used, with its bottom layer set to be Gaussian. We train the model  layer by layer using the method developed in the paper. The learning rate is set to be $10^{-3}$ . After training, samples were drawn from this deep models using Gibbs sampling. Specifically, we first use Gibbs sampler to draw samples from the distribution constituted by the first two top layers. Then, we pass down the samples through the hierarchical graphical model, and treat the samples at the bottom layer as the generated images. From Figure \ref{face_fig}, we can see that the generated faces looks like true face images.
	
\end{document}